\documentclass{article}

\PassOptionsToPackage{numbers, compress}{natbib}
\usepackage[preprint]{arxiv}

\usepackage[utf8]{inputenc} % allow utf-8 input
\usepackage[T1]{fontenc}    % use 8-bit T1 fonts
\usepackage{hyperref}       % hyperlinks
\usepackage{url}            % simple URL typesetting
\usepackage{booktabs}       % professional-quality tables
\usepackage{amsfonts}       % blackboard math symbols
\usepackage{nicefrac}       % compact symbols for 1/2, etc.
\usepackage{microtype}      % microtypography
\usepackage{xcolor}         % colors
\usepackage{graphicx}
\usepackage{caption}
\usepackage{subcaption}
\usepackage{hyperref}
\usepackage[numbers]{natbib} 
\usepackage{authblk}
\usepackage{amsmath}
\usepackage{ulem}

\hypersetup{
    unicode=false,          % non-Latin characters in Acrobat’s bookmarks
    pdftoolbar=true,        % show Acrobat’s toolbar?
    pdfmenubar=true,        % show Acrobat’s menu?
    pdffitwindow=false,     % window fit to page when opened
    pdfstartview={FitH},    % fits the width of the page to the window
    pdftitle={},    % title
    pdfauthor={},     % author
    pdfsubject={},   % subject of the document
    pdfcreator={},   % creator of the document
    pdfproducer={}, % producer of the document
    pdfkeywords={wildfire forecasting} {S2S} {deep learning} {teleconnections} {transformers}, % list of keywords
    pdfnewwindow=true,      % links in new window
    colorlinks=true,       % false: boxed links; true: colored links
    linkcolor=blue,          % color of internal links (change box color with linkbordercolor)
    citecolor=blue,        % color of links to bibliography
    filecolor=blue,      % color of file links
    urlcolor=blue           % color of external links
}

\title{TeleViT1.0: Teleconnection-aware Vision Transformers for Subseasonal to Seasonal Wildfire Pattern Forecasts}

\author[1, 2, *]{Ioannis Prapas}
\author[1]{Nikolaos Papadopoulos}
\author[1, 3]{Nikolaos-Ioannis Bountos}
\author[3]{Dimitrios Michail}
\author[2]{Gustau Camps-Valls}
\author[1]{Ioannis Papoutsis}

\affil[1]{Orion Lab, National Technical University of Athens}
\affil[2]{Image Processing Laboratory (IPL), Universitat de Val\`encia}
\affil[3]{Department of Informatics and Telematics, Harokopio University of Athens}
\begin{document}

\maketitle

\begin{abstract}

Forecasting wildfires weeks to months in advance is difficult, yet crucial for planning fuel treatments and allocating resources. While short-term predictions typically rely on local weather conditions, long-term forecasting requires accounting for the Earth’s interconnectedness, including global patterns and teleconnections. We introduce TeleViT, a Teleconnection-aware Vision Transformer that integrates (i) fine-scale local fire drivers, (ii) coarsened global fields, and (iii) teleconnection indices. This multi-scale fusion is achieved through an asymmetric tokenization strategy that produces heterogeneous tokens processed jointly by a transformer encoder, followed by a decoder that preserves spatial structure by mapping local tokens to their corresponding prediction patches.

Using the global SeasFire dataset (2001–2021; 8-day resolution), TeleViT improves AUPRC performance over U-Net++, ViT, and climatology across all lead times, including horizons up to four months. At zero lead, TeleViT with indices and global inputs reaches AUPRC 0.630 (ViT 0.617, U-Net 0.620); at 16×8-day lead ($\sim$ 4 months), TeleViT variants using global input maintain $\sim$ 0.601–0.603 (ViT 0.582, U-Net 0.578), while surpassing the climatology (0.572) at all lead times. Regional results show the highest skill in seasonally consistent fire regimes, such as African savannas, and lower skill in boreal and arid regions. Attention and attribution analyses indicate that predictions rely mainly on local tokens, with global fields and indices contributing coarse contextual information. These findings suggest that architectures explicitly encoding large-scale Earth-system context can extend wildfire predictability on subseasonal-to-seasonal timescales.

\end{abstract}

\section{Introduction}

% it is important to forecast wildfires at the S2S scales
Wildfires are a growing global concern, driven by climate change and expanding human pressures, leading to more frequent and severe fire seasons \citep{jones_global_2022}. In recent years, record-breaking wildfires across multiple continents have led to catastrophic impacts, from widespread loss of life and property to prolonged smoke pollution affecting millions of people \citep{cunningham2025climate, kolden2025wildfires}. These escalating risks underscore the urgent need for improved preparedness rather than response. If fire managers and policymakers could anticipate dangerous wildfires weeks to months in advance, they would be better equipped to mitigate adverse effects, with improved fuel management and optimized resource allocation \citep{national_academies_of_sciences_next_2016}. 

% the current operational status
Many operational systems for wildfire forecasting leverage weather-driven indices, such as the Fire Weather Index (FWI), losing predictive skill beyond the medium range scale (1-14 days). For Subseasonal-to-Seasonal (S2S) timescales the European Forest Fire Information System (EFFIS) is using expected temperature and precipitation anomalies \citep{noauthor_effis_nodate-1}. However, these do not directly translate to operational decisions that need targeted operational variables, instead of expected weather anomalies \citep{national_academies_of_sciences_next_2016}. Despite the current status, most pressing preparedness decisions depend on reliable information at S2S timescales. 

% teleconnections modulating burned area
For weather, the S2S forecasting horizon has long been considered as a predictability desert, where neither weather models nor climate projections offer robust skill. However, for S2S wildfire forecasts, one promising avenue is to leverage teleconnections —the well-established climate oscillations that drive weather patterns globally —and slow-varying ``memory'' effects in the Earth system. In fact, it has been shown that these slow, large-scale Earth system processes heavily modulate global wildfires \citep{cardil_climate_2023, kuhn-regnier_importance_2021}. For instance, extreme wildfires in Siberia have been linked to preceding Arctic Oscillation patterns and previous-year soil moisture anomalies \citep{kim_extensive_2020}. Globally, teleconnections have been shown to modulate 52.9\% of the burned area \citep{cardil_climate_2023}, while antecedent vegetation and drought conditions can substantially improve burned area models \citep{kuhn-regnier_importance_2021}.

% promise of deep learning
Harnessing these sources of predictability in practice, however, is not straightforward as different biomes exhibit distinct teleconnection-fire relationships on concurrent and time-lagged bases, underscoring complex mechanisms by which climate modes and memory effects affect fire activity. Deep Learning (DL) methods offer a promising avenue for learning intricate spatiotemporal interactions in the Earth system, learning complex patterns from data and capturing the dynamic interactions among various fire drivers and the resulting burned areas \citep{reichstein_deep_2019}. In recent years, we have witnessed significant advances in DL-based medium-range (1-14 days) forecasts \citep{rasp_weatherbench_2024}, where DL models have been shown to surpass the skill of state-of-the-art Numerical Weather Prediction (NWP) models. While still early, this type of models are starting to show skill at the S2S scales, particularly with the use of large ensembles, which are produced significantly faster and more cheaply from DL models than NWP-based models. GenCast \citep{price2023gencast}, a diffusion-based model initially developed for medium-range forecasting, has demonstrated strong performance at the S2S \citep{antonio2025seasonal}. Similarly, Fuxi-S2S \citep{chen_machine_2024} advances ensemble forecasting by introducing flow-dependent perturbations into the model’s hidden states, yielding more robust predictions. Complementing these advancements, \citet{bommer2025deep} demonstrated that explicitly incorporating relevant climate fields, such as the North Atlantic Oscillation, can enhance S2S forecast skill of European winter weather.  

% dl for s2s wildfire forecasting
Several studies have successfully applied DL to wildfire forecasting tasks \citep{jain_review_2020}, demonstrating superior predictive performance compared to traditional methods for short-term forecasting \citep{kondylatos_wildfire_2022}. Here, we focus on a limited number of studies that have dealt with wildfire forecasting at S2S scales. \citet{torres-vazquez_enhancing_2025} shows how a simple drought indicator can provide skillful predictions of burned area anomalies a month before the start of the target fire season in about $68\%$ of the burnable area. At the same time, the performance is enhanced with seasonal NWP predictions.  \citet{chen_how_2016} use simple autoregressive statistical methods to combine volume pressure deficit values with oceanic indicators. \citet{yu_quantifying_2020} uses a statistical pre-processing to identify the most prominent oceanic indicators modulating burned area in Africa and then uses the findings to select input features for tree-based machine learning models. AttentionFire \citep{li_attentionfire_v10_2023} uses an attention-enhanced recurrent neural network architecture that considers temporal context and shows improvements at longer horizons when adding information from climatic indices. \citet{michail_firecastnet_2025} uses a graph architecture inspired by GraphCast \citep{lam_graphcast_2022}, aiming to jointly capture temporal dependencies and spatial interactions jointly, showing increased model performance with long input time-series. 

% contributions
Motivated by the growing evidence that long-range wildfire predictability arises from large-scale global processes such as teleconnections and memory effects, this work presents Teleconnection-aware Vision Transformer (TeleViT), a deep learning model specifically designed to embed these known sources of predictability into its architecture. An initial study \citep{prapas_televit_2023} hinted that TeleViT can effectively integrate local fire driver variables with coarsened global variables and time series of teleconnection indices to enhance global burned area pattern forecasting. In this study, we substantially advance this line of research by refining and presenting the architecture in detail, while also evaluating tokenization strategies for different inputs. Beyond global forecasting skill, we provide a comprehensive assessment of regional performance and examine the model attention and attribution to understand the relative contribution of the different inputs.
More specifically, this work makes the following contributions:

\begin{itemize}
    \item We propose TeleViT, a transformer model that explicitly models both local and non-local Earth system variables, enabling it to achieve improved performance in burned-area pattern forecasting up to 4 months in advance. 
    \item We rigorously evaluate the models and analyze the performance across different fire regimes. Additionally, inspect the model's attention and explainability to identify how much focus is placed on the different inputs (local, global, teleconnection indices) and which variables are most important.
    \item We provide open data, source code, and a toolkit that allows to inspect the models and their predictions (see \ref{sec:code-availability}).\end{itemize}
% The remainder of this paper is organized as follows: Section \ref{sec:televit} introduces the TeleViT architecture and the main methods of the study. Section \ref{sec:experimental-setup} presents our experimental setup, followed by the results and discussion in Section \ref{sec:results}. Finally, \ref{sec:limitations} concludes the study, discussing the main findings of this work, its limitations and potential future research directions.

\section{Teleconnection-aware Vision Transformer (TeleViT)}
\label{sec:televit}

\begin{figure*}[t]
    \centering
    \includegraphics[width=\linewidth]{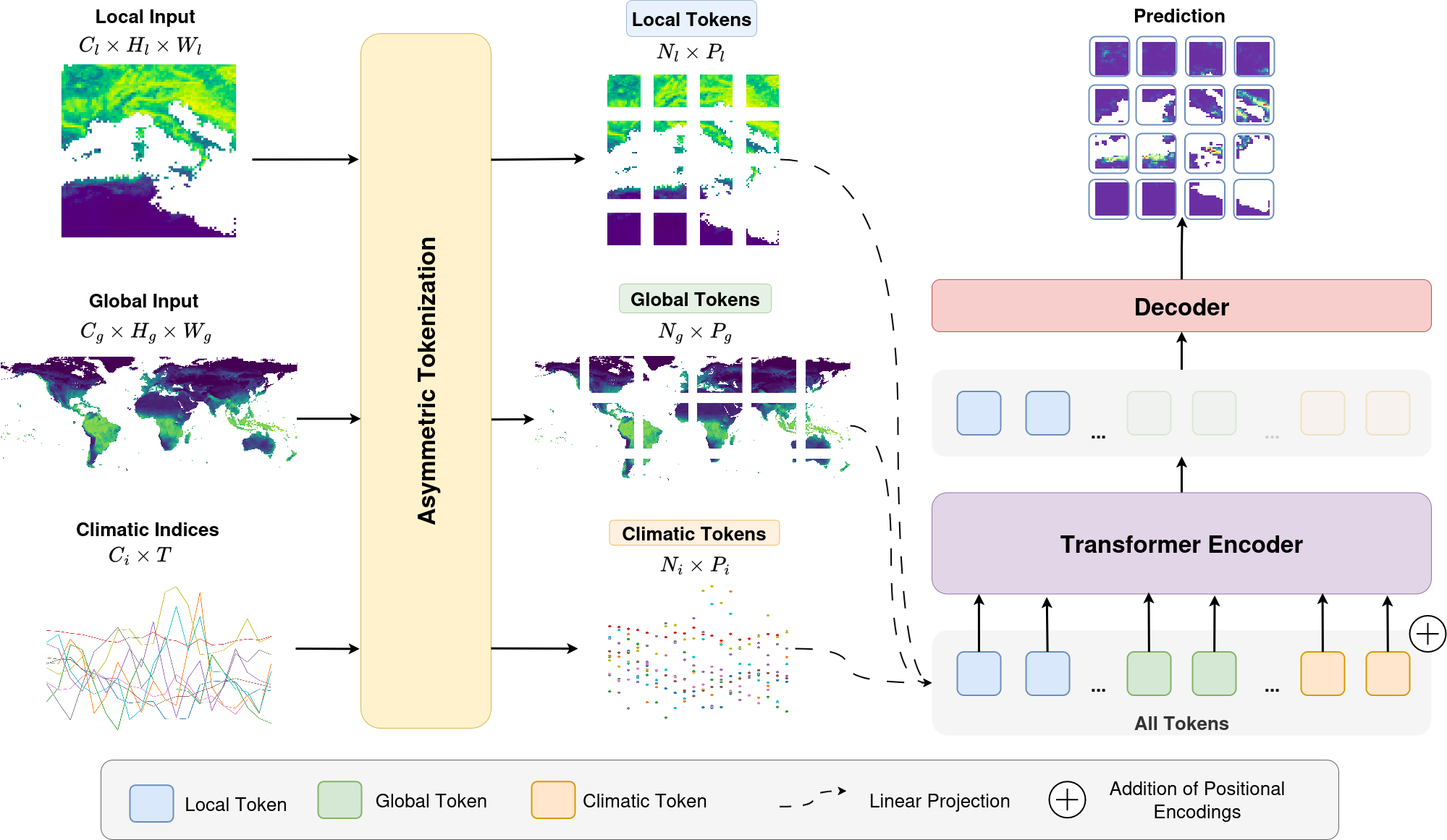}
    \caption{Architecture of the TeleViT model. The model processes three types of inputs (local, global, and indices) through an asymmetric tokenization strategy, followed by a transformer encoder and linear decoder to produce predictions.}
    \label{fig:pipeline}
\end{figure*}

The TeleViT architecture, as shown in Figure \ref{fig:pipeline}, is designed to capture both local- and global-scale interactions among the input variables. 
TeleViT leverages an asymmetric tokenization to split the different inputs into tokens, then treats them equally, as in the transformer encoder of ViT \citep{dosovitskiy_image_2021}. To produce the final output, a linear projection decoder is applied to the tokens representing the local input. This is presented in detail in the following subsections, along with the methods we use to inspect what the model has learned in section \ref{sec:model-inspection}.

\subsection{Architecture Overview}

TeleViT processes three types of inputs as follows:

\begin{itemize}
    \item Local input $x^{(l)} \in \mathbb{R}^{C_l \times H_l \times W_l}$: Finer-resolution data for the target region. Captures fine-grained local conditions, including soil moisture, vegetation state, and recent weather patterns.
    \item Global input $x^{(g)} \in \mathbb{R}^{C_g \times H_g \times W_g}$: Coarse-resolution global data. Represents broader patterns that can reveal teleconnections, such as large-scale temperature and precipitation, ocean surface temperatures, and atmospheric circulation.
    \item Indices input $x^{(i)} \in \mathbb{R}^{C_i \times T}$: Time series of teleconnection indices (e.g., ENSO, NAO, PDO) that quantify the strength and phase of known teleconnection patterns, providing explicit information about long-range connections.
\end{itemize}
where $C_k$ represents the number of features, $H_k$ and $W_k$ are the spatial dimensions, and $T$ is the temporal length for indices.

The key innovation in TeleViT lies in its asymmetric tokenization strategy, which allows it to process each input type differently based on its inherent characteristics and scale. For each input type $k \in \{l, g, i\}$, we define a tokenization function $\mathcal{T}_k: * \rightarrow \mathbb{R}^{N_k \times P_k}$ that splits the input into $N_k$ tokens of size $P_k$. Each token is projected to a common embedding dimension $D$ using input-specific linear projections $f_k: \mathbb{R}^{N_k \times P_k} \rightarrow \mathbb{R}^{N_k \times D}, \quad k \in \{l, g, i\}$. The resulting token sequence is augmented with the addition of learnable positional encodings $E_{pos} \in \mathbb{R}^{(N_l + N_g + N_i) \times D}$, which is a common way to address the permutation invariance of the transformer and give each token an identity. While in this case we use three types of input, we could actually leverage any input if provided with an appropriate tokenization. 

A standard Transformer encoder processes the token sequence as in the original ViT architecture \citep{dosovitskiy_image_2021} with $K$ layers and $A$ attention heads per layer. The final prediction is produced by applying a linear decoder to each local token. This decoder projects the embedding dimension $D$ into the target spatial resolution $H_l \times W_l$ of its corresponding local patch. The outputs of all local decoders are then assembled to reconstruct the full prediction map. In contrast, the initial TeleViT design relied on a single classification token and a large decoder to generate the final output, which effectively collapsed the spatial structure of the input. In the revised approach, each local token retains responsibility for its own spatial region, ensuring that the model preserves spatial coherence while still leveraging global context through attention.

\subsection{Inspection of the model}
\label{sec:model-inspection}

We use two main ways to inspect what the model has learned. First, by examining the attention weights presented in section \ref{sec:attention-weights}, we can identify salient tokens and the relative contributions of the different token types. Second, with the integrated gradients (\ref{sec:integrated-gradients}), we can see the most important variables for a target location. 

\subsubsection{Leveraging the attention weights}
\label{sec:attention-weights}

The attention mechanism produces attention maps of size $K \times A \times N \times N$, where $K$ is the number of layers, $A$ is the number of attention heads, and $N = N_l + N_g + N_i$ is the total number of tokens (sum of local, global, and index tokens). We average the heads and focus on the last layer's attention weights after computing the attention roll-out \citep{abnar_quantifying_2020}, which effectively represents the cumulative attention flow through all layers:

$$
\bar{\mathbf{A}}= \prod_{l=1}^{L} \frac{A^{(l)} + I}{\sum_j (A^{(l)}_{ij} + I_{ij})}
$$

where $\mathbf{A}_l$ represents the attention matrix at layer $l$. The resulting attention matrix $\bar{\mathbf{A}} \in \mathbb{R}^{N \times N}$ can be partitioned into blocks corresponding to different token types:

$$
    \bar{\mathbf{A}} = \begin{bmatrix}
    \mathbf{A}_{ll} & \mathbf{A}_{lg} & \mathbf{A}_{li} \\
    \mathbf{A}_{gl} & \mathbf{A}_{gg} & \mathbf{A}_{gi} \\
    \mathbf{A}_{il} & \mathbf{A}_{ig} & \mathbf{A}_{ii}
    \end{bmatrix}
$$

where subscripts $l$, $g$, and $i$ denote local, global, and indices tokens respectively. Given that the decoder uses only the local tokens of the last layer for the prediction, we focus on analyzing the following attention blocks: i) local-to-local attention ($\mathbf{A}_{ll} \in \mathbb{R}^{N_l \times N_l}$), ii) local-to-global attention ($\mathbf{A}_{lg} \in \mathbb{R}^{N_l \times N_g}$), iii) local-to-indices attention ($\mathbf{A}_{li} \in \mathbb{R}^{N_l \times N_i}$).

\subsubsection{Integrated Gradients}
\label{sec:integrated-gradients}

Given an input image \( \mathbf{x} \in \mathbb{R}^n \) and a baseline \( \mathbf{x}' \in \mathbb{R}^n \), the attribution for each input feature \( i \) is computed as:

$$
\mathrm{IG}_i(\mathbf{x}) = (x_i - x_i') \times \int_0^1 \frac{\partial F\big(\mathbf{x}' + \alpha (\mathbf{x} - \mathbf{x}') \big)}{\partial x_i} \, d\alpha
$$

where $ F: \mathbb{R}^n \to \mathbb{R}^m$ is the sum of the computed score for the positive class, and $\alpha \in [0,1] $ interpolates between the baseline and the input. We set a zero baseline, which is a common choice when the inputs are normalized during training. Given the multi-dimensional inputs and outputs, we perform a coarse-scale inspection to identify the most important variable in each local patch. 

\subsection{Key Features}

The architecture supports rich interactions across and within datasets. The attention mechanism allows information to flow between data sources and across scales, capturing both global patterns and local details. Independent input tokenization enables flexible input handling, while each local token retains spatial context and draws on global signals to make predictions.

This per-token decoder design is handy for analyzing how different input types influence predictions. Since each local token contributes directly, attention weights reveal the relevance of regional, global, and index tokens. With appropriate tokenization, TeleViT can be adapted to accommodate new data inputs with different dimensions. 

% Figure~\ref{fig:attn_multires} illustrates these attention-based interactions, highlighting both intra-dataset (blue) and inter-dataset (orange) connections. This interpretable framework not only models complex Earth system dynamics but is also easily adaptable to new data types and timescales.

% \begin{figure}[htbp]
% \centering
% \includegraphics[width=6cm]{figs/attention_matrix.png}
% \caption{Depiction of intra-dataset (blue) and inter-dataset interactions (orange) in an attention matrix for a sequence of six tokens with an equal number of tokens for each data source.}
% \label{fig:attn_multires}
% \end{figure}

\section{Experimental Setup}
\label{sec:experimental-setup}

\subsection{Machine Learning Task}

 We follow a similar setup to \citep{prapas_deep_2022, prapas_televit_2023}, who initially presented the task of burned area pattern forecasting as a segmentation task. Our task is to predict the future burned area pattern, i.e., the presence of burned areas at a target location at a future time step. That is, given an input sample $x_t$ at time $t$, our goal is to develop a model $f(\cdot)$ that predicts the presence of burned areas $y_{t+}$ at a future time $t+h$, where $h$ represents the forecasting horizon:

$$
    \hat{y}_{t+h} = f(x_t), \quad h \in \mathbb{N}
$$

\subsection{Dataset and Preprocessing}

\begin{table}[htbp]
\centering
\small
\renewcommand{\arraystretch}{0.9}
\caption{Input and target variables used from the SeasFire cube for all settings. Same variables used for both local and global views. Variables marked with an asterisk (*) were log-transformed as a preprocessing step.}
\label{tab:datacube-variables}
\begin{tabular}{@{}p{0.55\linewidth}p{0.4\linewidth}@{}}
\toprule
\textbf{Local/Global Variables} & \textbf{Climatic Indices} \\
\midrule
Mean sea level pressure & Western Pacific Index (WP) \\
Total precipitation * & Pacific North American Index (PNA) \\
Vapor Pressure Deficit & North Atlantic Oscillation (NAO) \\
Sea Surface Temperature & Southern Oscillation Index (SOI) \\
Mean Temperature at 2 meters & Arctic Oscillation (AO) \\
Surface solar radiation downwards & Pacific Decadal Oscillation (PDO) \\
Volumetric soil water level 1 & Eastern Asia/Western Russia (EA) \\
Land Surface Temperature at day & East-North Pacific Oscillation (EPO) \\
Normalized Difference Vegetation Index & Ni\~no 3.4 Anomaly (NINo34) \\
Population density * & Bivariate ENSO (CENSO) \\
Cosine of longitude & \\
Sine of longitude  & \\
Cosine of latitude & \\
Sine of latitude  & \\
\midrule
\multicolumn{2}{l}{\textbf{Target Variable}} \\
\midrule
Binarized burned area from GWIS & \\
\bottomrule
\end{tabular}
\renewcommand{\arraystretch}{1}
\end{table}

We conduct our experiments on the SeasFire cube \citep{karasante2025seasfire}, a spatio-temporal dataset, suitable for subseasonal to seasonal wildfire forecasting. It contains 21 years of data (2001-2021) at a global scale, with a temporal resolution of 8 days and a spatial resolution of $0.25^{\circ}$. The cube includes a diverse range of fire drivers, combining atmospheric, vegetation, and anthropogenic variables with climate indices, as well as target variables related to wildfires, namely burned areas, fire radiative power, and wildfire-related emissions. Note that while the datacube is organized in 8-day time steps, throughout the text we interchangeably use the term ``week'' (and related terms such as ``weekly”) for simplicity and clarity. When we refer to weeks in this text, we mean the 8-day intervals in the SeasFire dataset.

Table \ref{tab:datacube-variables} shows which variables are used from the SeasFire cube, along with the pre-processing applied to each variable. Total precipitation and population are log-transformed using $log(1+x)$ to reduce skewness while preserving zeros.. For more details on the variables, the reader is referred to the cited dataset descriptor \citep{karasante2025seasfire}. The selection of the variables follows \citet{prapas_televit_2023} and \citet{michail_firecastnet_2025}, who have previously used SeasFire for S2S wildfire pattern forecasting. 

To train our data-driven models, we construct input–target pairs $(x,y)$. For baseline models, only local inputs $x^{(l)}$ are used, whereas TeleViT models additionally require global inputs $x^{(g)}$ and index inputs $x^{(i)}$. The local input is obtained by extracting $80 \times 80$ patches. Since the spatial domain is represented at a resolution of $0.25^{\circ}$ (corresponding to $1440 \times 720$ grid cells), it can be divided into $18 \times 9 = 162$ local patches. The global input is constructed by coarsening the grid to a $1^{\circ}$ resolution, reducing its size by a factor of 16, resulting in a $360 \times 180$ grid. For both local and global inputs, 10 fire-driver variables are extracted from the cube (Table \ref{tab:datacube-variables}). Additionally, we compute a \textit{global positional encoding}, defined as the sine and cosine of longitude and latitude; this is distinct from the transformer’s own positional encoding. For each sample, we extract 10 indices corresponding to the 10 months preceding time $t$. The target $y$ is defined as the burned area at time $t+h$ within the region corresponding to the local input, where $h$ denotes the lead-time forecasting horizon. Consequently, each sample consists of four components: the local input $x^{(l)} \in \mathbb{R}^{14 \times 80 \times 80}$, the global input $x^{(g)} \in \mathbb{R}^{14 \times 360 \times 180}$, the indices input $x^{(i)} \in \mathbb{R}^{10 \times 10}$, and the target $y \in \mathbb{R}^{1 \times 80 \times 80}$. As a preprocessing step, all inputs are standardized. 
 
\subsection{Models, Baselines and Evaluation}

We assess the following models: i) a U-Net++, which uses only local input $x_l$, ii) a simple ViT, which uses only local input $x_l$, iii) $\mathrm{TeleViT}_{i}$, which uses indices $x_i$ along with local input $x_l$, iv) $\mathrm{TeleViT}_{g}$, which uses only global input $x_g$ along with local input $x_l$, and v) $\mathrm{TeleViT}_{i,g}$, which uses both indices $x_i$ and global input $x_g$ along with the local input $x_l$. The performance of the models is examined in several forecasting horizons $h \in \{0, 1, 2, 4, 8, 16\} \times$ 8-day steps. A model that predicts at the maximum forecasting horizon, i.e., 16$\times$8 days in advance, learns to predict the burned-area pattern for a particular 8-day period approximately 4 months in advance. A different model is trained for each $h$. 

We use the burned-area pattern climatology as a statistical baseline. In this work, the term \textit{Climatology} is used loosely to refer to the historical likelihood of a cell being burned during a given 8-day period, computed across all training years. Formally, it quantifies the empirical frequency of fire occurrence at a specific 
location and week of the year, without conditioning on any additional predictors.

The performance of the different models is evaluated using the Area Under the Precision-Recall Curve (AUPRC). By ignoring true negatives and focusing solely on precision and recall for the positive class, AUPRC is sensitive to how well a model captures burned areas, even at the expense of recall. The train, validation, and test split is time-based, using years 2003 - 2017 for training, 2018 for validation and 2019 for testing. The first two years of the dataset, 2001 and 2002, are excluded from training because the consistency of burned-area observations improves after the launch of MODIS Aqua in May 2002. Patches that contain exclusively ocean are excluded. 

After an analysis of the effect of the different hyperparameters on the validation performance presented in the Appendix \ref{sec:appendix-hparams}, we find that smaller architectures are preferred for the task and that moderate tokenization is enough for the local and global input, while the indices input needs fine-grained tokenization to prevent performance degradation. The encoders of both ViT and TeleViT consist of $K=8$ layers, with $A=8$ attention heads each, an embedding dimension $D=768$ and an MLP dimension of size $1536$. For the asymmetric tokenization, we set $P_l=(1, 16, 16)$, $P_g=(1, 30, 30)$ and $P_i=1$. This means that i) the local input $x_l$ of size $(14, 80, 80)$ is tokenized spatially in $5\times5$ number of tokens with size $16\times16$ ii) the global input $x_g$ of size $(14, 360, 180)$ is tokenized spatially in $12\times6$ number of tokens with size $30\times30$, and iii) the teleconnection indices input $x_i$ of size $(10, 10)$ is tokenized in $10\times10$ number of tokens.

Models are trained for 30 epochs. We use the cross-entropy loss and the Adam optimizer to train the models. For the U-Net++ model, the initial learning rate is set to 0.001, while for the Transformer models, it is set to 0.0001. Cosine annealing is used for learning rate scheduling, with a warmup period of 5\% of the total steps. The model with the lowest validation loss is used for testing. 

\section{Results and Discussion}
\label{sec:results}

\subsection{Input Types Comparison}

\begin{figure}[htbp]
    \centering
    \includegraphics[width=\linewidth]{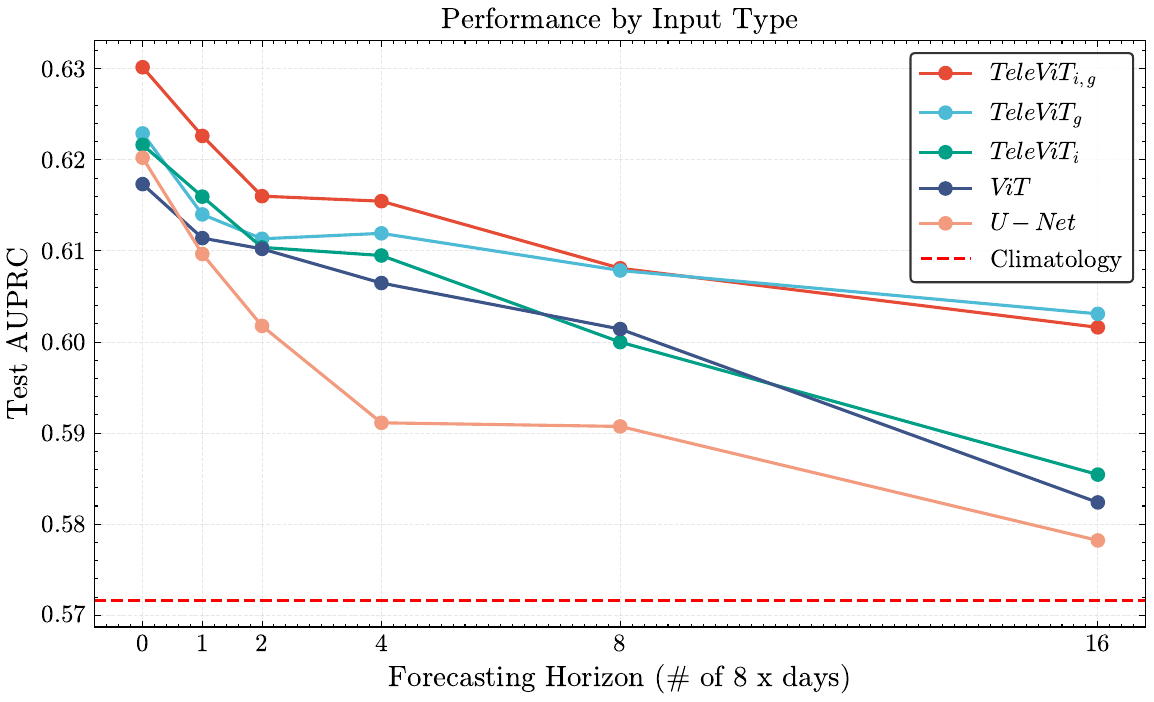}
    \caption{Performance comparison across different input configurations and lead times. The plot shows how different combinations of input types affect model performance, with TeleViT$_{i,g}$ consistently outperforming other variants across all forecasting horizons.}
    \label{fig:input_types_comparison}
\end{figure}

We conduct a comprehensive comparison of different input types to understand their relative contributions to model performance. Figure \ref{fig:input_types_comparison} shows the performance comparison across different input configurations and lead times, evaluating the effectiveness of adding global and indices inputs in all combinations. In general, all the models exhibit the expected decline in performance as the forecasting horizon increases. The decline, however, is less steep for teleconnection-aware models that use global-scale inputs ($\mathrm{TeleViT}_{i,g}$, $\mathrm{TeleViT}_{g}$). ViT proves to be a stronger baseline, outperforming U-Net++ for all forecasting horizons except $h=0$. $\mathrm{TeleViT}_{g}$ achieves a comparable performance to ViT for short forecasting windows of up to 4$\times$8 days, while it shows greater robustness to the increase of the forecasting window. With the inclusion of teleconnection indices, $\mathrm{TeleViT}_{i}$ achieves comparable performance to the traditional ViT that has only access to the local input. Interestingly, $\mathrm{TeleViT}_{i,g}$ shows the best performance for the first forecasting horizons (0, 1, 2, 4 weeks in advance), indicating a potential synergistic effect between teleconnection indices and global-scale representations.

Although TeleViT models exhibit a larger performance gap relative to the Climatology baseline, all DL models appear more skillful than the baseline, even when predicting 16 weeks in advance. While initial work \citep{prapas_televit_2023} used the continuous version of the climatology, i.e., the mean seasonal cycle of burned area, we find this version more suitable, as it captures the historical likelihood of a cell being burned. This is because the AUPRC, which is used to assess skill, is calculated by setting different thresholds for the predictor variable. A standard threshold based on global burned area undermines skill at the regional scale, since other regions exhibit different fire magnitudes. 

\begin{figure*}[htbp]
    \centering
    \includegraphics[width=\linewidth]{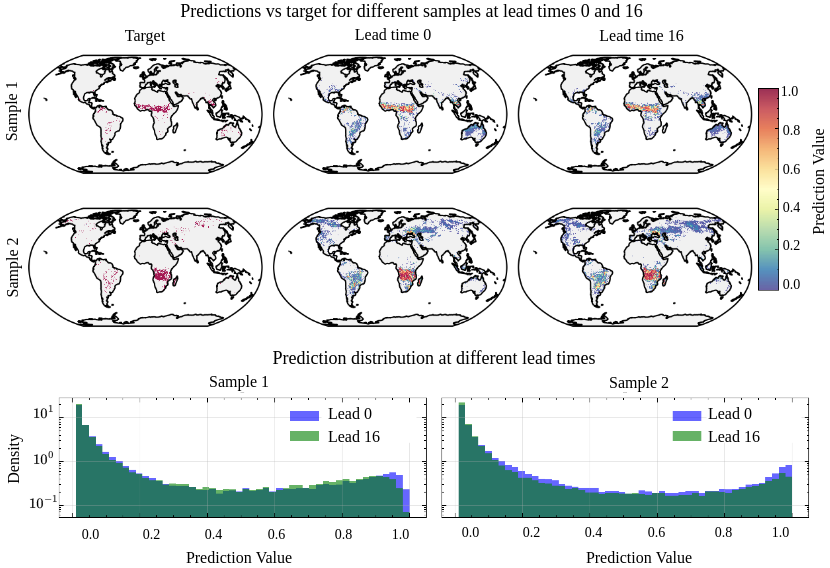}
    \caption{Sample global predictions (week 0 and week 25) of the test year versus the target for the best model, predicting at a lead forecasting horizon of 0 and 16 $\times$ 8-day steps. Values below 0.01 are masked. Confidence is determined as the softmax score of the positive prediction. In the bottom, we see the distributions of the prediction scores for the positive class for the two sample global prediction maps.}
    \label{fig:danger_maps}
\end{figure*}

The fire danger maps (Figure \ref{fig:danger_maps}) show how the predictions compare to the target across different lead time horizons, with high agreement.  Notably, the models capture major shifts in fire activity across Australia, North America, and Eastern Europe, as well as a change in fire patterns from the northern to the southern African savanna.  At first glance, it is not evident how the predictions differ at the two extremes of the forecasting horizons ($h=0$ and $h=16$), and their performance differences are subtle. For sample 2, it is evident that, with the large forecasting horizon, the series are slightly smoother. We can verify this by looking at the histograms of the predictions, which show greater density in the high (low) confidence predictions for the h=0 (h=16) model.

\subsection{Regional Performance Analysis}

\begin{figure*}[htbp]
    \centering
    \includegraphics[width=\linewidth]{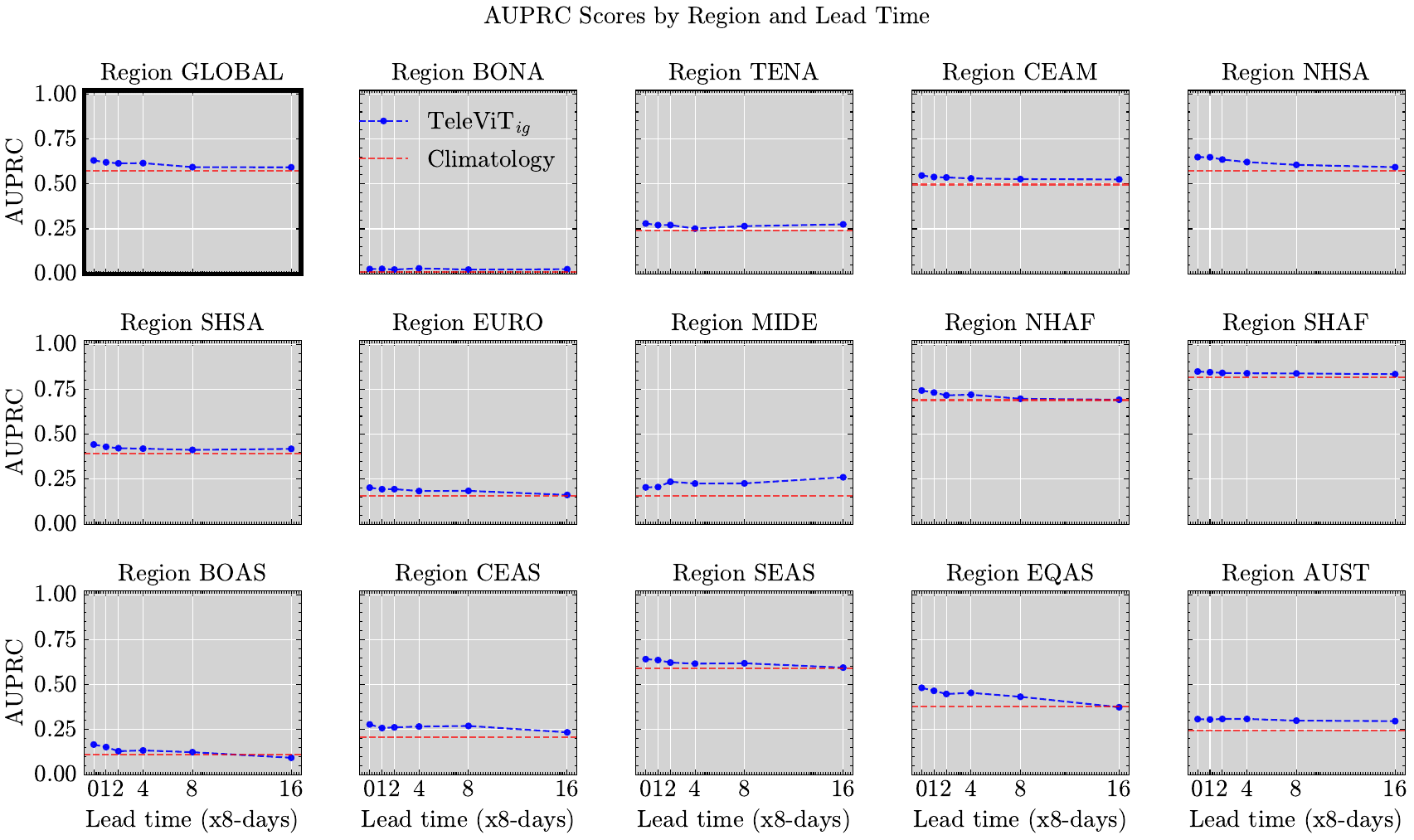}
    \caption{AUPRC scores of the Televit$_{ig}$ model for all GFED regions and lead times. The performance of the Climatology baseline is shown as a red dashed horizontal line.}
    \label{fig:auprc_by_region_and_lead_time}
\end{figure*}

To better understand how the $\mathrm{TeleViT}_{ig}$ model performs across different global regions, we analyze its performance according to the Global Fire Emissions Database (GFED) regional classification. Figure \ref{fig:auprc_by_region_and_lead_time} shows the AUPRC scores for each GFED region across different forecasting horizons. The performance of the prediction model is always very close to the Climatology baseline. We see that across all areas, the model is skillful (outperforming the statistical baseline), and, except for MIDE, which is an outlier, its performance converges towards that of the baseline as the forecasting horizon increases. 

In terms of absolute performance values, each GFED region’s fire regime reflects a mix of the region's climate, vegetation, and human activity \citep{jones_global_2022}. African savannas (NHAF, SHAF) lead with the highest scores, thanks to their frequent and highly seasonal fires. Tropical regions in Latin America (NHSA, SHSA, CEAM) and Southeast and Equatorial Asia (SEAS, EQAS) show moderate performance, with AUPRC values ranging from 0.4 to 0.6. Temperate Mediterranean regions (TENA, EURO) and arid zones (MIDE, CEAS) have lower AUPRCs (0.2-0.4). Their fires are sporadic with complex or fuel-limited regimes that weaken the predictability. Equatorial Asia (EQAS) also has a low AUPRC, with fires there producing weak, irregular patterns. Australia exhibits lower AUPRC values ($\simeq$ 0.2), reflecting weak spatial seasonality in burned-area patterns. Finally, wildfires in boreal regions (BONA, BOAS) are temperature-driven in environments rich in fuel, making spatial patterns less predictable and resulting in a substantially low AUPRC ($<$ 0.2).

Overall, having been trained on past burned area patterns, the model struggles to predict areas that do not exhibit spatially consistent fire patterns. To improve performance in complex or highly variable fire regimes, we would need additional region-specific refinement and potentially consider the size of burned areas. 

\subsection{Model Inspection}

\begin{figure*}[htbp]
    \centering
    \includegraphics[width=\linewidth]{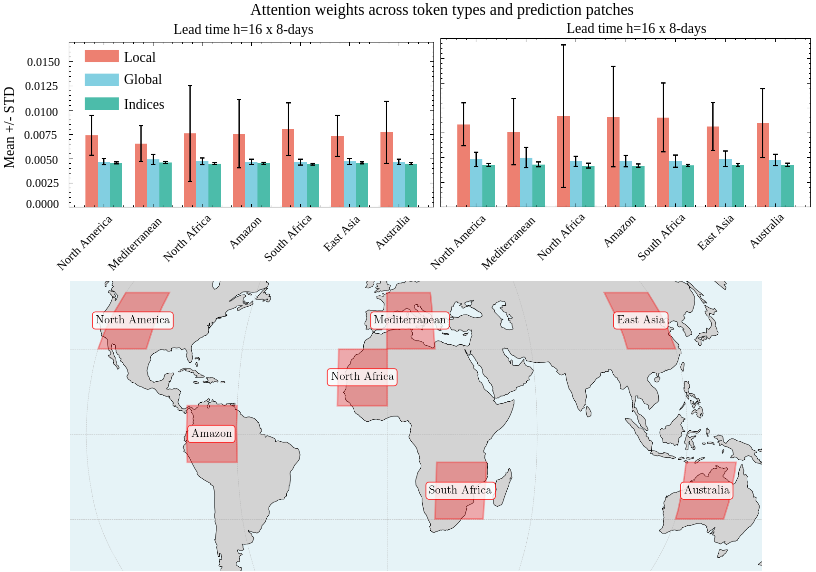}
    \caption{Bar charts of mean attention weights with standard deviations (error bars) for each token type (local, global, indices) across prediction patches around the world. The left chart shows results for a forecasting horizon of $h=0$ $\times$ 8-days, while the right chart shows results for $h=16$ $\times$ 8-days. The bottom map indicates the geographical locations of the selected patches.}
    \label{fig:attention_values}
\end{figure*}

Following the approach described in Section \ref{sec:attention-weights}, we can investigate how much the model relies on each token type in different regions for the final prediction. We analyze the mean attention weights for different token types across regions to understand how the model integrates local, global, and index information during prediction. Figure \ref{fig:attention_values} summarizes the relative contribution of each input type for the test set. Local tokens receive higher attention weights and also exhibit a high standard deviation. We find that the standard deviation of regional-to-local attention is exceptionally high in regions with diverse vegetation patterns, ranging from savannas to rainforests, such as the Amazon and North and South Africa. This indicates the model is making selective, discriminative choices about which specific pieces of information to focus on. The high variance means some tokens receive much more attention than others, suggesting the model can identify and prioritize specific, fine-grained details. Attention to global tokens is lower, with a low standard deviation that differs slightly across locations. The indices tokens generally attract the least attention across regions. When attention weights are similar across many inputs, the model is essentially averaging information rather than making selective choices. From this, we infer that the model derives general context from global-scale inputs rather than fine-grained details.

\begin{figure*}[htbp]
    \centering
    \includegraphics[width=\linewidth]{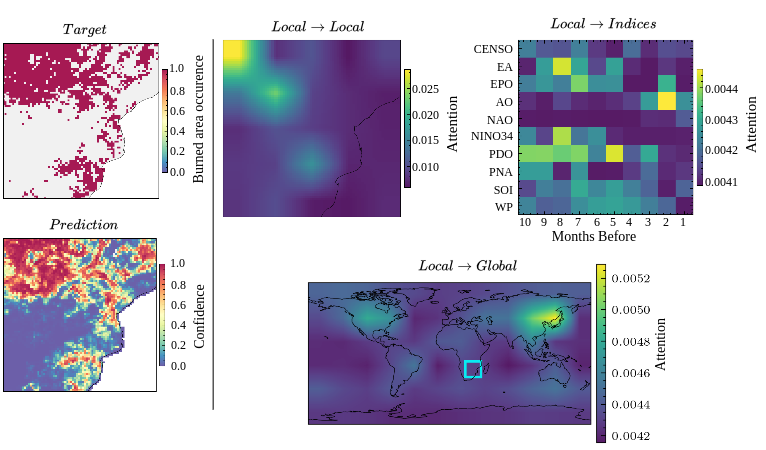}
    \caption{Left side: Prediction vs target (first row), local-to-local attention (middle left), local-to-indices attention (middle right), local-to-global attention (bottom)}
    \label{fig:attention_maps}
\end{figure*}

To complement the analysis, we further visualize the different attention pathways alongside the target and predicted burned area maps for a representative sample (Figure \ref{fig:attention_maps}), focusing on the blue-highlighted region in southeastern Africa. The spatial distribution of salient local-to-local attention closely aligns with the prediction, indicating that local tokens capture fine-scale features critical for the forecast. In contrast, the local-to-global and local-to-indices attention maps exhibit lower magnitudes and more diffuse patterns. Importantly, these visualizations should not be interpreted as direct causal evidence. For example, attention to global regions such as East Asia or to indices like the Arctic Oscillation at a two-month lag cannot be taken to imply a mechanistic link to the South African burned area. Since their attention weights have low values and are very similar in magnitude, the search for salient indices and global patches is not justified; the model forms a blurry view of the global state while relying primarily on localized information for predictive accuracy. This type of analysis illustrates how attention-based inspection can reveal the model's internal strategies, shedding light on the relative contributions of local and non-local drivers.

\begin{figure*}[htbp]
    \centering
    \includegraphics[width=\linewidth]{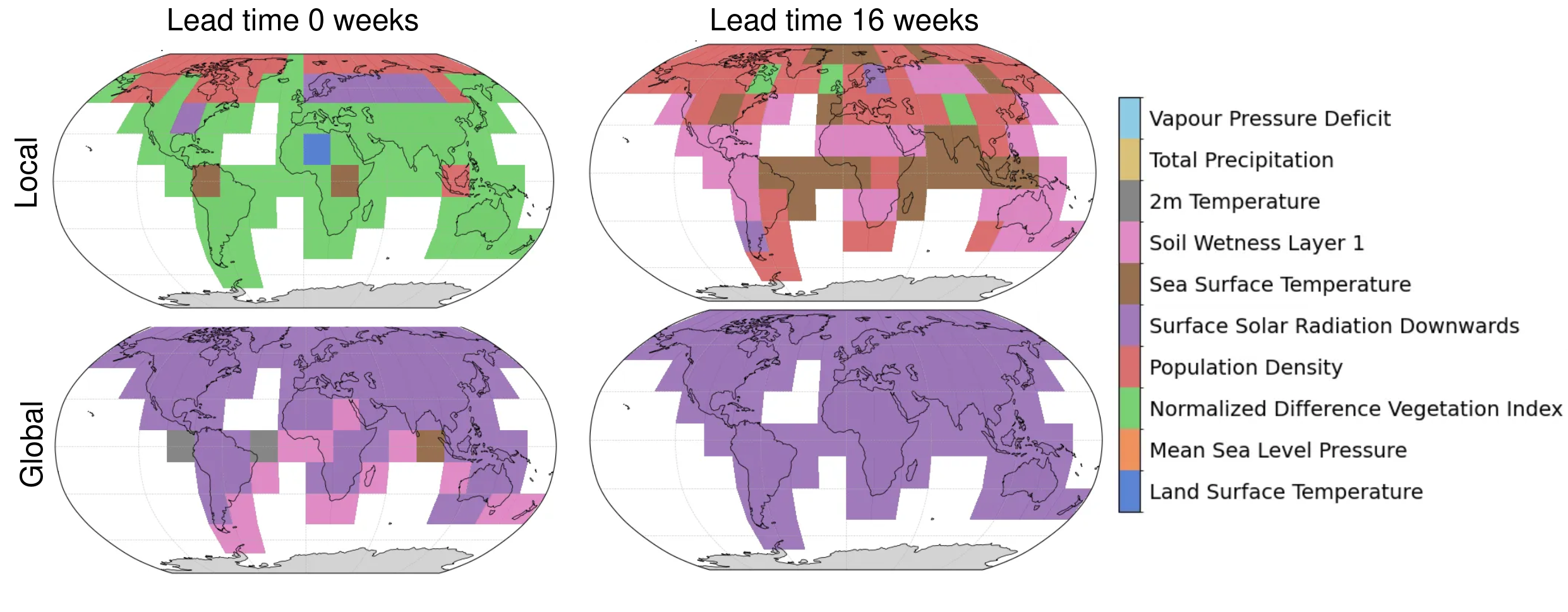}
    \caption{Most important local (top row), vs global (bottom row) variable per patch for lead time=0 (left), 16 (right) $\times$ 8-days.}
    \label{fig:integrated-gradients}
\end{figure*}

To identify the most important local and global variables for each patch, we use the integrated gradients approach (Section \ref{sec:integrated-gradients}). Figure \ref{fig:integrated-gradients} shows the most important local and global variables for each patch for two forecasting horizons ($h=0$ and $h=16$ $\times$ 8-days). In the short term, the vegetation index appears to dominate the importance of local inputs. In the long term, the most important variables are population density, soil moisture, and sea surface temperature, with some spatial continuity, indicating that nearby regions have similar fire regimes. For the global input, solar radiation is the model's primary focus, suggesting that an essential aspect of the global input is to provide seasonality information. For the indices, which are provided to the model as individual tokens, the use of integrated gradients is not very meaningful as the values would be influenced mainly by individual index values relative to the zero baseline.

To further inspect the model's attention to different inputs, we provide a Huggingface application (redacted for double-blind review), where a user can visualize the predictions, attentions, and integrated gradient attributions at different time steps and forecasting horizons for the entire test dataset (year 2019).

\section{Conclusion}
\label{sec:limitations}

Predictions on S2S scales can profit from the predictability offered by large-scale phenomena in the Earth system, such as memory effects and teleconnections. Especially in wildfire forecasting, it is well known that antecedent conditions and atmospheric teleconnections drive many wildfires. In this work, we proposed TeleViT, a teleconnection-driven transformer that combines local and non-local inputs, such as coarsened global variables and time series of teleconnection indices, to predict burned-area patterns up to 4 months in advance. The model improves performance against models that do not leverage such information, with more persistent gains as the forecasting horizon increases. Analyzing performance across regions, we find that the model performs best in regions with spatially consistent fire patterns. By examining the model's attention weights, we see that it learns to focus more on local tokens, independent of the forecasting horizon. Global and index tokens receive less attention, but they nevertheless help improve performance by providing global context. 

Despite the contributions of this study, several limitations remain, indicating directions for future work to build on the present findings.
Though the TeleViT model achieves the best performance when using a single time-step as input, \citet{michail_firecastnet_2025} has demonstrated that improved results are possible by incorporating temporal information. For TeleViT, this could involve developing temporal tokenization strategies and positional encodings for the different input types, potentially leading to better representation of the complex spatio-temporal relationships of the other variables. In that case, it would be beneficial to also look into more efficient attention schemes that address its quadratic complexity \citep{katharopoulos_transformers_2020}. While the inclusion of global coarsened variables enables superior performance, incorporating teleconnection indices did not significantly improve forecasts and increased model complexity by adding a large number of tokens. 
Analysis of variable importance revealed that the model ignored known teleconnections and instead learned weak or non-meaningful associations with several teleconnection indices. Instead, future work can learn to respect known causal relationships between teleconnections, constraining the attention to make plausible associations. In that direction, \citet{zhao2024causal} explored how to improve the generalization between regions respecting a causal structure between teleconnection indices and the burned areas. Given that the impacts of teleconnections often emerge only above certain intensity thresholds, we could also use categorical values, as in \citet{bommer_deep_2025}, rather than absolute intensities. Additionally, at S2S scales it is worth exploring methods that produce probabilistic forecasts in the direction of GenCast \citep{price2023gencast} and Fuxi-S2S \citep{chen_machine_2024}. 

To conclude, this work demonstrated that deep learning models explicitly designed to capture Earth system interactions can be beneficial for long-term forecasting. The combination of local information, climatic indices, and global views of the Earth holds significant potential for various applications. While this architecture was initially developed for burned area pattern forecasting, it would be worthwhile to investigate how it can be used to predict burned area anomalies or other Earth system variables with relevant characteristics, such as vegetation stress or soil moisture.

\section*{Funding Statement}
This work is part of the SeasFire project, which focuses on ``Earth System Deep Learning for Seasonal Fire Forecasting,'' and is funded by the European Space Agency (ESA) within the ESA Future EO-1 Science for Society Call.

\section*{Code and Data Availability}
\label{sec:code-availability}
This work leverages the SeasFire dataset, version v.0.4 \citep{karasante2025seasfire}, which is publicly available in analysis-ready format \citep{alonso_seasfire_2022}. Our codebase is available at \url{[https://github.com/orion-AI-Lab/televit1.0]} under an open-source license, and includes tools for loading and preprocessing the dataset, as well as training the various TeleViT versions and baseline models. Users can modify most aspects of the model training, as well as input variables and tokenization schemes directly through the configuration file. In addition, we provide a Huggingface application at \url{[https://huggingface.co/spaces/iprapas/televit-xai]}, under an open-source license, enabling interactive visualization of attention maps, predictions, and integrated-gradients attributions across time steps and forecasting horizons for all the test set.

\appendix

\section{Main results in table form}

\begin{table*}[h]
\caption{This table shows the performance of the models in terms of AUPRC for the test year (2019) for the different lead time horizons of 0, 1, 2, 4, 8, and 16 $\times$ 8 days. Best scores are bold, and second-best scores are underlined.}
\label{tab:results-table}
\begin{tabular}{lcccccc}
\toprule
                        & \multicolumn{6}{c}{\textbf{Lead Time (\# of 8-days)}}                                                     \\ 
\textbf{Model} & 0 & 1 & 2 & 4 & 8 & 16 \\ \midrule
Climatology     & \multicolumn{6}{c}{0.5716}                                                                                 \\ \midrule
ViT            & 0.6173          & 0.6114          & 0.6102          & 0.6065          & 0.6014          & 0.5824          \\
U-Net++            & 0.6202          & 0.6097          & 0.6018          & 0.5911          & 0.5907          & 0.5782          \\ \midrule
TeleViT$_{i}$   & 0.6217          & \uline{0.6160}          & 0.6104          & 0.6095          & 0.6000          & 0.5854          \\
TeleViT$_{g}$   & \uline{0.6229}          & 0.6140          & \uline{0.6113}          & \uline{0.6119}          & \uline{0.6079}          & \textbf{0.6031} \\
TeleViT$_{i,g}$ & \textbf{0.6302} & \textbf{0.6226} & \textbf{0.6160} & \textbf{0.6155} & \textbf{0.6081} & \uline{0.6016}          \\ 
\bottomrule

\end{tabular}
\end{table*}

\section{Impact of key parameters}
\label{sec:appendix-hparams}

\begin{figure*}[htbp]
    \centering
    \includegraphics[width=0.9\linewidth]{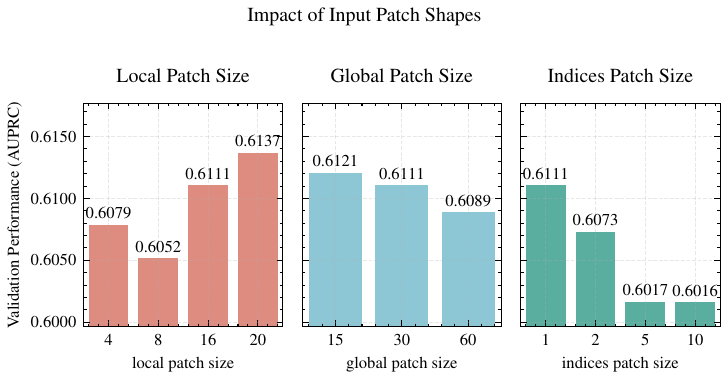}
    \caption{Impact of different patch configurations on model performance. Left: Local patch size analysis showing optimal performance at 16$\times$16. Middle: Global patch size analysis showing best performance with 30$\times$30 patches. Right: Indices patch-size analysis showing best performance with the smallest size of 1×1.}
    \label{fig:ablation_input_shapes}
\end{figure*}

Here, we investigated some key parameters. More specifically, Figure \ref{fig:ablation_input_shapes} demonstrates the impact of different patch sizes for the other input types(local, global, and indices shows the results of these experiments, and Figure \ref{fig:ablation_individual_parameters} shows the impact of modifying key aspects of the architecture (MLP dimension, number of heads, depth, positional variables). We start from a complete TeleViT model trained to predict a lead time of 16 $\times$ 8-days in advance, as we are more interested in the parameters that affect long-term performance. The initial parameters are a 16$\times$16 local patch size, a 30$\times$30 global patch size, a 1$\times$1 indices patch size, 12 heads, 12 layers, and an MLP dimension of 1536. Each parameter is modified separately to assess its individual effect, and we measure validation AUPRC.

\subsection{Impact of tokenization strategy}

Key findings from the patch size analysis:
\begin{itemize}
    \item \textbf{Local Patch Size}: Tested with four sizes: 4, 8, 16, and 20. Performance peaks at a local patch size of 20, shows comparable performance between 4 and 8, and increases with larger patch sizes. Surprisingly, a finer patch size does not help improve performance.
    
    \item \textbf{Global Patch Size}: Tested with three sizes: 15, 30, and 60. Performance seems to drop slightly with larger patch sizes, while a patch size of 30 seems sufficient. This indicates that moderate global patch sizes are enough for capturing global context.
    
    \item \textbf{Indices Patch Size}: Tested with four different sizes: 1, 2, 5, 10 applied in the time dimension. Shows a sharp decreasing trend as patch size increases. Best performance with the smallest size of 1, with a significant drop in performance for larger sizes (2, 5, 10). This suggests that climate indices are most effective when considered at individual time points.
\end{itemize}

\subsection{Impact of transformer architecture parameters and ablation of positional variables}

\begin{figure*}[htbp]
    \centering
    \includegraphics[width=\linewidth]{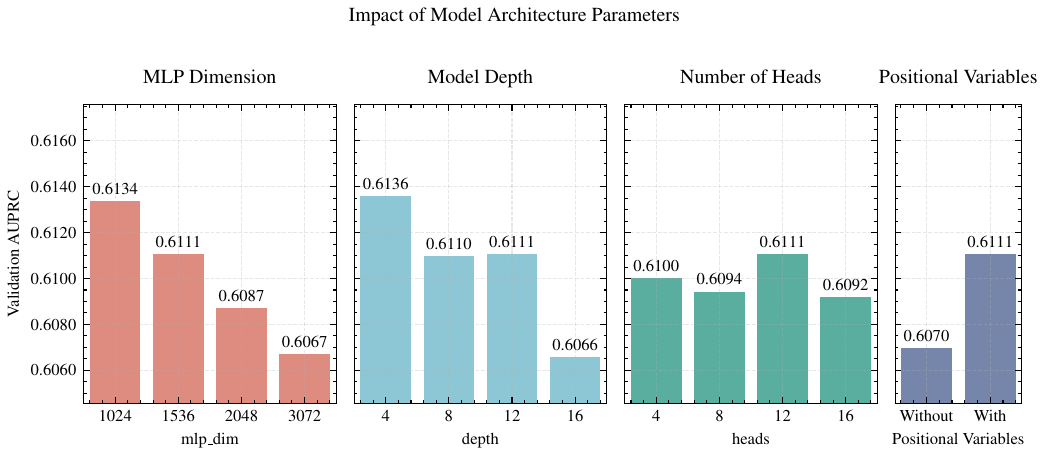}
    \caption{Impact of model architecture parameters on performance. Left: MLP dimension analysis. Middle-left: Model depth analysis. Middle-right: Number of heads analysis. Right: Positional variables ablation.}
    \label{fig:ablation_individual_parameters}
\end{figure*}

Here, we investigate the impact of different transformer architecture parameters. Figure \ref{fig:ablation_individual_parameters} shows the results of the experiments where 

Key findings from the parameter analysis:
\begin{itemize}
    \item \textbf{MLP Dimension}: Tested three sizes: 1536, 2048, and 3072. Best performance achieved with 1536, showing a generally decreasing trend with larger dimensions. This suggests that smaller MLP dimensions are sufficient for the task, while larger dimensions may lead to unstable training prone to overfitting.
    
    \item \textbf{Model Depth}: Evaluated depths of 4, 8, 12, and 16 layers. Best performance with the shallowest model, with a clear decreasing trend as depth increases. This indicates that the task doesn't require very deep architectures, and simpler models may be more effective and computationally efficient.
    
    \item \textbf{Number of Heads}: Tested 4, 8, 12, and 16 attention heads. Best performance with 12 heads, while the effect of changing the number of heads is not significant, achieving comparable performance with four heads as well.
    
    \item \textbf{Positional Variables}: Small benefit clear from including positional variables.
\end{itemize}

\bibliographystyle{unsrtnat}
\bibliography{references.bib}

\end{document}